\documentclass{IEEEtran}
\usepackage{cite}
\usepackage{amsmath,amssymb,amsfonts}
\usepackage{graphicx}
\usepackage{textcomp}
\usepackage{multirow}
\usepackage{hyperref}
\usepackage{multicol}
\usepackage{url}
\usepackage{booktabs}
\usepackage{makecell}
\usepackage{color}
\usepackage{bbding}
\usepackage{algorithm}
\usepackage{algpseudocode}
\def\BibTeX{{\rm B\kern-.05em{\sc i\kern-.025em b}\kern-.08em
		T\kern-.1667em\lower.7ex\hbox{E}\kern-.125emX}}
\ifCLASSINFOpdf
\else
\fi
\begin{document}
	\title{Graph Learning-Driven Multi-Vessel Association: Fusing Multimodal Data for Maritime Intelligence}
	\author{Yuxu Lu, Kaisen Yang, Dong Yang, Haifeng Ding, Jinxian Weng, and Ryan Wen Liu
		\thanks{The work described in this paper is supported by the Research Grants Council of the Hong Kong Special Administrative Region, China (Grant Number. PolyU 15201722). (Corresponding author: Dong Yang and Ryan Wen Liu)}
		\thanks{Yuxu Lu, Kaisen Yang, and Dong Yang are with the Department of Logistics and Maritime Studies, the Hong Kong Polytechnic University, Hong Kong 999077, and also with the Maritime Data and Sustainable Development Centre, the Hong Kong Polytechnic University, Hong Kong 999077. (e-mail: yuxulouis.lu@connect.polyu.hk, kaisen.yang@connect.polyu.hk and dong.yang@polyu.edu.hk)}
		\thanks{Haifeng Ding and Jinxian Weng are with the College of Transport and Communications, Shanghai Maritime University, Shanghai 201306, China. (e-mail: dinghaifeng0008@stu.shmtu.edu.cn and jxweng@shmtu.edu.cn)}
		\thanks{Ryan Wen Liu is with the School of Navigation, Wuhan University of Technology, Wuhan 430063, China, and also with the State Key Laboratory of Maritime Technology and Safety, Wuhan 430063, China. (e-mail: wenliu@whut.edu.cn)}
	}
\maketitle
\begin{abstract}
    Ensuring maritime safety and optimizing traffic management in increasingly crowded and complex waterways require effective waterway monitoring. However, current methods struggle with challenges arising from multimodal data, such as dimensional disparities, mismatched target counts, vessel scale variations, occlusions, and asynchronous data streams from systems like the automatic identification system (AIS) and closed-circuit television (CCTV). Traditional multi-target association methods often struggle with these complexities, particularly in densely trafficked waterways. To overcome these issues, we propose a graph learning-driven multi-vessel association (GMvA) method tailored for maritime multimodal data fusion. By integrating AIS and CCTV data, GMvA leverages time series learning and graph neural networks to capture the spatiotemporal features of vessel trajectories effectively. To enhance feature representation, the proposed method incorporates temporal graph attention and spatiotemporal attention, effectively capturing both local and global vessel interactions. Furthermore, a multi-layer perceptron-based uncertainty fusion module computes robust similarity scores, and the Hungarian algorithm is adopted to ensure globally consistent and accurate target matching. Extensive experiments on real-world maritime datasets confirm that GMvA delivers superior accuracy and robustness in multi-target association, outperforming existing methods even in challenging scenarios with high vessel density and incomplete or unevenly distributed AIS and CCTV data.
\end{abstract}
\section{Introduction}\label{sec:intr}
\begin{figure}[t]
        \centering
        \setlength{\abovecaptionskip}{0.cm}
        \includegraphics[width=1.00\linewidth]{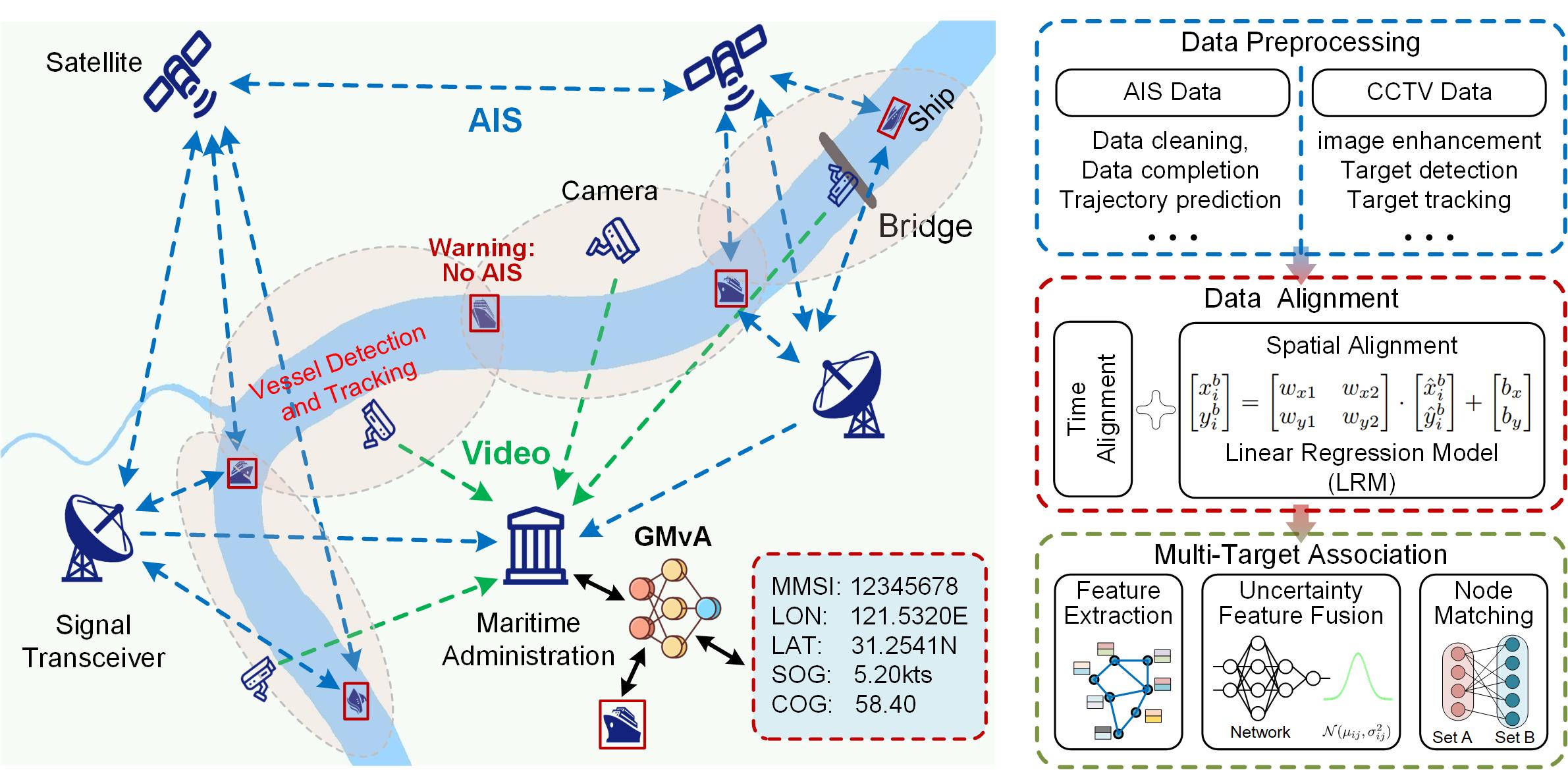}
        \caption{Illustration of maritime intelligent transportation systems (MITS) for complex waterways. The integration of AIS data and video enables reliable multi-target association, ensuring accurate vessel tracking, identity verification, and behavior monitoring. Our method maintains seamless surveillance continuity, even in the presence of system failures or data inconsistencies.}
        \label{figure01}
    \end{figure}

    \IEEEPARstart{M}{aritime} intelligent transportation systems (MITS) achieve a full-chain regulatory loop of "\textit{perception-decision-response}" through multimodal data fusion in complex waterways \cite{chen2022reliable}. Its core value is reflected in two dimensions: real-time anomaly behavior warnings (e.g., sudden course deviations, traffic lane violations) \cite{hu2022intelligent} and post-incident multimodal evidence tracing \cite{radovic2012post}. To meet these demands, as shown in Fig. \ref{figure01}, MITS leverage advanced sensing technologies by integrating two complementary data sources: the automatic identification system (AIS) and closed-circuit television (CCTV) \cite{guo2023asynchronous}. AIS provides dynamic vessel information, such as position, course, speed, and identity \cite{yang2019big}, while CCTV captures real-time, high-fidelity optical and infrared visual data, enabling precise analysis of vessel behaviors and spatial relationships within monitored zones \cite{liu2021enhanced,forti2022next}. The fusion of maritime multi-modal data sources allows MITS to detect and track vessel anomalies more effectively, identify unsafe navigation behaviors, and support early recognition of potential collision risks \cite{zhang2025systems}. Moreover, CCTV imagery provides real-time critical evidence for post-incident investigations, helping authorities reconstruct events and analyze the contributing factors to maritime accidents. However, accurately associating AIS data with CCTV data in high-traffic environments remains challenging, specifically,
    \begin{itemize}
        \item 	\textit{Temporal and spatial misalignment}: Differences in temporal resolution, spatial coverage, and vessel dynamics between AIS and CCTV make alignment challenging, increasing the possibility of association errors.
        \item 	\textit{Inconsistent data quality}: The instability of AIS signals and the variability of CCTV video quality reduce the reliability of data, especially in cases of occlusion, bad weather, or equipment problems.
    \end{itemize}
    \begin{table*}[t]
        \centering
        \footnotesize
        \setlength{\tabcolsep}{8.50pt}
        \caption{Comparison of point-, trajectory-, and graph-based maritime multimodal data fusion methods.}
        \renewcommand{\arraystretch}{1.2}
        \begin{tabular}{p{3.0cm}|p{6.5cm}p{6.5cm}}
            \hline
            \textbf{Fusion Types} & \textbf{Advantages} & \textbf{Disadvantages} \\ \hline\hline
            Point-based \cite{lu2021fusion,zhao2025multi}&
            Fast and simple. \newline
            Effective for sparse, well-separated targets. &
            Easily affected by noise and missing data. \newline
            Poor performance in dense scenarios. \\ \hline
            Trajectory-based \cite{wu2022new,guo2023asynchronous}&
            Captures temporal dynamics for better robustness. \newline
            Handles motion patterns like speed and heading. &
            Struggles with overlapping or incomplete trajectories. \newline
            Sensitive to abrupt motion changes. \\ \hline
            Graph-based \cite{qu2023improving,yang2022multi}&
            Robust modeling capabilities for complex scenarios. \newline
            Effective for global association tasks. &
            High computational cost. \newline
            Relies on accurate graph construction and clean data. \\ \hline
        \end{tabular}\label{tab:matching_comparison}
    \end{table*}

    Currently, multimodal maritime data fusion is achieved through three main processes, i.e., data preprocessing, data alignment and multi-target association \cite{selvaraj2023machine}. During data preprocessing, visual data undergoes image enhancement, target detection, and multi-target tracking to extract vessel behavior features \cite{tian2018detection,zhang2023unsupervised}. Simultaneously, AIS data is processed through noise filtering, missing value imputation, and trajectory prediction to enhance data integrity \cite{bernabe2023detecting}, thereby providing reliable input for the alignment and association of multimodal data. Data alignment involves time synchronization to match the timestamps of AIS and CCTV, as well as spatial alignment to transform AIS coordinates into a perspective consistent with video frames. During the data association phase, as shown in Table \ref{tab:matching_comparison}, the proposed methods in the past can be broadly classified into three categories: point-based, trajectory-based, and graph-based methods.
    
    Point-based methods primarily rely on proximity in location and time to preliminarily associate data points from different sources \cite{lu2021fusion,zhao2025multi}. Trajectory-based methods further leverage the motion patterns of vessels, enhancing the robustness of matching by predicting and associating continuous motion trajectories \cite{guo2023asynchronous,lei2024association}. Nevertheless, point- and trajectory-based methods are prone to instability under conditions of fluctuating data quality and increased complexity in navigation environments, as they overly depend on single association rules. In contrast, graph-based methods \cite{yang2022multi} can build interaction networks between vessels, integrating spatial, temporal, and behavioral patterns to effectively address issues of multi-target interference and ambiguity. For example, by analyzing the relative motion relationships and interaction behaviors between vessels, graph-based methods can more accurately identify individual targets within a group of vessels. Furthermore, graph matching methods combined with deep learning and graph neural networks (GNN) \cite{yang2022multi} can automatically extract more complex spatiotemporal features, thereby further enhancing the accuracy and robustness of the associations.
    
    This work mainly focuses on \textit{the challenge of multi-target association} in multimodal maritime data and proposes a novel graph-based multi-vessel association (GMvA) method. The GMvA constructs dynamic graphs for each modality, where nodes represent target observations. Edges encode temporal connections across consecutive frames, capturing temporal evolution patterns. Our GMvA integrates temporal graph attention (TGA) layer and spatial-temporal attention (STA) block to effectively learn discriminative feature representations from CCTV and AIS trajectories. These features are then fused through a dedicated multilayer perceptron-based uncertainty fusion module (MLP-UMF), which computes similarity scores between potential matches by seamlessly incorporating both appearance and geometric constraints. To ensure globally consistent assignments, a Hungarian algorithm-based matching optimization module is suggested, capable of handling missing targets and outliers. GMvA empowers the model to learn optimal feature representations directly from the data, eliminating the need for hand-crafted features or heuristic matching rules. Extensive experimental results validate its effectiveness and robustness. The main contributions of this work are summarized as follows
    \begin{itemize}
        \item  We propose a novel GMvA method that integrates spatial and temporal feature extraction through TGA Layer and STA block, thereby achieving modeling of vessel motion relationships from multiple perspectives.
        \item  We propose an effective MLP-UMF that combines learned node representations with geometric compatibility measures, providing robust matching performance even under challenging conditions with measurement uncertainties.
        \item  Our GMvA demonstrates exceptional adaptability to varying traffic densities and data source inconsistencies, achieving significant performance improvements over competitive methods in complex waterway scenarios.
    \end{itemize}

    The rest of this paper is organized as follows. Section \ref{sec:relatedwork} reviews related work on maritime data fusion. In Section \ref{sec:definition}, we initially define the research problem. Section \ref{sec:ourmethod} proposes the multi-vessel association method. Extensive experiments are implemented to evaluate the performance of our method in Section \ref{sec:experiments}. Conclusion is given in Section \ref{sec:conclusions}.	
\section{Related Work}\label{sec:relatedwork}
   Maritime multimodal data fusion significantly enhances MITS by providing more comprehensive, accurate, and timely vessel monitoring and anomaly detection capabilities \cite{liu2024efficient}. Existing methods can be broadly classified into three categories: point-based, trajectory-based, and graph-based methods.
\subsection{Point-based Fusion}
    Point-based matching establishes correspondences between multi-source observations based on spatiotemporal proximity. For example, Habtemariam \textit{et al.} \cite{habtemariam2015measurement} adopted the joint probabilistic data association framework to directly fuse radar measurements with AIS messages. Similarly, Gaglione \textit{et al.} \cite{gaglione2018belief} proposed a belief propagation-based AIS/radar fusion method, constructing a factor graph within a Bayesian framework to solve target state estimation and data association problems. Visual-based fusion methods have further expanded the capabilities of AIS matching. Monocular camera systems can estimate the distance and azimuth angle between vessels and the camera, with the resulting data fused with AIS to achieve vessel identification. To enhance accuracy, dead reckoning methods are often integrated to predict short-term vessel trajectories, reducing errors caused by visual distance estimation \cite{lu2021fusion}. Multi-region mapping and feature similarity measures have also been developed to improve robustness in AIS-visual data matching \cite{ding2024robust}. In addition, distance-speed correction methods, based on the pinhole imaging principle, help mitigate data fluctuations from visual feature drift or vessel occlusion \cite{ding2024real}. Recent advancements in image-based methods have focused on improving the alignment of AIS data with visual observations. For example, homography-based coordinate transformations coupled with object detection models \cite{gulsoylu2024image}. Similarly, spatial alignment models convert AIS data into pixel coordinates and employ hybrid constraints (e.g., field of view, position, and heading) to match visual and AIS targets more effectively \cite{zhao2025multi}. While these point-level methods are computationally efficient and straightforward, they face limitations in complex scenarios with noisy data, asynchronous timestamps, or high vessel density. In such cases, the spatiotemporal proximity of multiple vessels can lead to ambiguous associations, requiring more robust and adaptable solutions. 
\subsection{Trajectory-based Fusion}
    Trajectory-based methods extend point-based methods by incorporating temporal dynamics, associating entire trajectories instead of discrete points. By analyzing motion patterns such as speed, heading, and trajectory shape, trajectory-based methods demonstrate increased robustness to noise, missing data, and inconsistencies. For example, Qin \textit{et al.} \cite{qin2018research} proposed an adaptive trajectory fusion method that uses historical data for weighted correction combined with nearest neighbor rules and distance-speed correlation methods, enabling smooth trajectory association. Similarly, Wu \textit{et al.} \cite{wu2022new} proposed a fuzzy evaluation-based trajectory association framework, integrating coarse and fine matching to align AIS, radar, and image target data. A-IPDA \cite{sun2023ais}, an AIS-assisted radar tracking method, efficiently updates target states and existence probabilities in occluded environments by integrating radar observations and AIS data through sequence processing. E-FastDTW \cite{guo2023asynchronous}, an enhanced dynamic time warping method, calculates trajectory similarity in asynchronous data and applies the Hungarian algorithm to achieve efficient AIS-video target matching. Jin \textit{et al.} \cite{jin2023radar} proposed a learning-based method that integrates trajectory and scene features, using three-dimensional Convolutional neural networks to extract spatiotemporal features. Other methods like evidence reasoning fusion \cite{xu2023integration} and dynamic classification \cite{lei2024association} attempt to improve fusion performance by integrating multi-dimensional features, trust weighting, and trajectory dynamics. However, trajectory-based methods face challenges in overlapping trajectories, abrupt motion changes, and inconsistencies caused by differing data sampling rates or incomplete trajectories.
\subsection{Graph-based Fusion}
    Graph-based matching uses graph structures to model relationships between multiple observations. Nodes represent vessels or data points, while edges capture spatial, temporal, or motion-based relationships. Graph-based methods excel in multi-target scenarios by capturing higher-order dependencies and global patterns. For example, Qu \textit{et al.} \cite{qu2023improving} constructed a weighted bipartite graph, using multidimensional features such as position, speed, and course as the basis for fusion. However, the bipartite graph matching method mainly matches based on static features, and it is challenging to dynamically update the weights to adapt to the real-time changing data distribution or feature relationships. GNN can simultaneously model node and edge features enables effective cross-modal feature fusion and global consistency modeling. For example, GNN has been effectively used to solve problems such as image-text matching \cite{chen2020uniter}, point cloud and graph structure matching \cite{fu2021robust}, knowledge graph alignment \cite{wang2018cross}, and dynamic scene registration \cite{yan2018spatial}. Their key advantages include the ability to achieve cross-modal feature fusion, global consistency modeling, and adaptability to a variety of task requirements, such as one-to-one matching, partial matching, and dynamic matching. For example, Yang \textit{et al.} \cite{yang2022multi} proposed a graph-matching-based network to transform AIS and radar track time-series data into graph distribution features. It leverages multi-scale point-level feature extraction and a graph neural network with self-cross attention to capture local and global features and achieves globally optimal matching through a differentiable optimal transport layer. GNN has shown significant potential in addressing cross-modal complex matching tasks. Consequently, efficiently leveraging graph structures to integrate multimodal maritime data and develop an accurate matching model remains a critical task.
    \begin{figure*}[t]
        \centering
        \setlength{\abovecaptionskip}{0.cm}
        \includegraphics[width=0.990\linewidth]{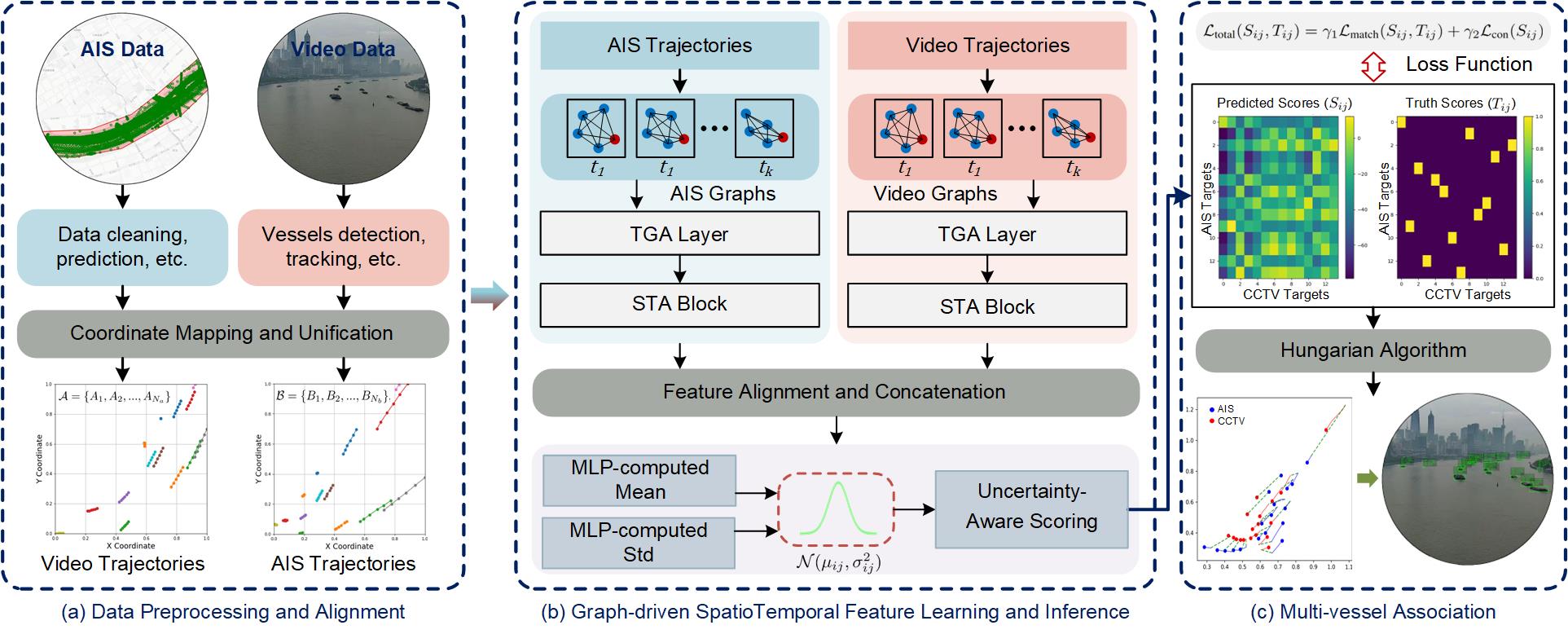}
        \caption{Overview of the proposed GMvA framework. At each timestamp, multimodal trajectories are structured into temporal graphs. High-dimensional node features are extracted via TGA layer and STA block, with feature normalization applied to enhance representation. After independently processing two distinct data streams, an MLP-NMF computes similarity scores between matching pairs, generating a cross-class similarity matrix. The Hungarian algorithm is then used to derive optimal matches from the matrix.}
        \label{figure:framework}
    \end{figure*}
\section{Initial Definition}\label{sec:definition}
\subsection{Problem Formulation}
    Consider two sets of vessel trajectory observations $\mathcal{A} = \{A_1, A_2, ..., A_{N_a}\}$ and $\mathcal{B} = \{B_1, B_2, ..., B_{N_b}\}$, where $\mathcal{A}$ is collected from AIS, and $\mathcal{B}$ is extracted from CCTV through vessel detection and tracking methods (more implementation details can be found in our previous work \cite{guo2023asynchronous}), followed by our suggested coordinate transformation method in subsection \ref{subsubsection:ucs}. $N_a$ and $N_b$ are the number of targets observed in AIS and CCTV respectively. Each trajectory is observed over a specific time window $T_w = \{t_1, t_2, \dots, t_k\}$, where $t_k$ represents a fixed timestamp within the time window. The observation at each time step captures the position state of a vessel, which can be given as
    \begin{equation}\label{eq:observation}
        A_i = (x_i, y_i), \quad B_j = (x_j, y_j),
    \end{equation}
    where $(x, y)$ represents the spatial coordinates in the unified normalized reference system. 
    
    Time is represented implicitly through the structure of the time window, ensuring that the observations at each time point correspond to a pre-defined list of timestamps. Missing observations for any timestamp within the time window are filled with zero. The extracted trajectories are stored as arrays with a shape of $[k, 2]$, where $k$ equals the number of timestamps in the time window. Finally, all trajectories are represented as a tensor with the shape $(N, k, 2)$, where $N = \operatorname{max}(N_a, N_b)$. To handle missing data and maintain consistent matrix dimensions, zero-padding is applied as needed. It can ensure that each trajectory point is aligned with the corresponding time window, even if certain timestamps are missing data.

    Our final objective is to learn a mapping function $\mathcal{F}_\text{GMvA}$ that generates an optimal matching matrix $\mathcal{M}$, i.e.,
    \begin{equation}
        \mathcal{M} = \mathcal{F}_\text{GMvA}(\mathcal{A}, \mathcal{B}; \Theta),
    \end{equation}
    where $\Theta$ represents the learnable parameters of our GMvA. The $\mathcal{M} \in \{0,1\}^{N \times N}$ must satisfy one-to-one matching constraints while maximizing the overall matching confidence.
    \begin{figure*}[t]
        \centering
        \setlength{\abovecaptionskip}{0.cm}
        \includegraphics[width=0.975\linewidth]{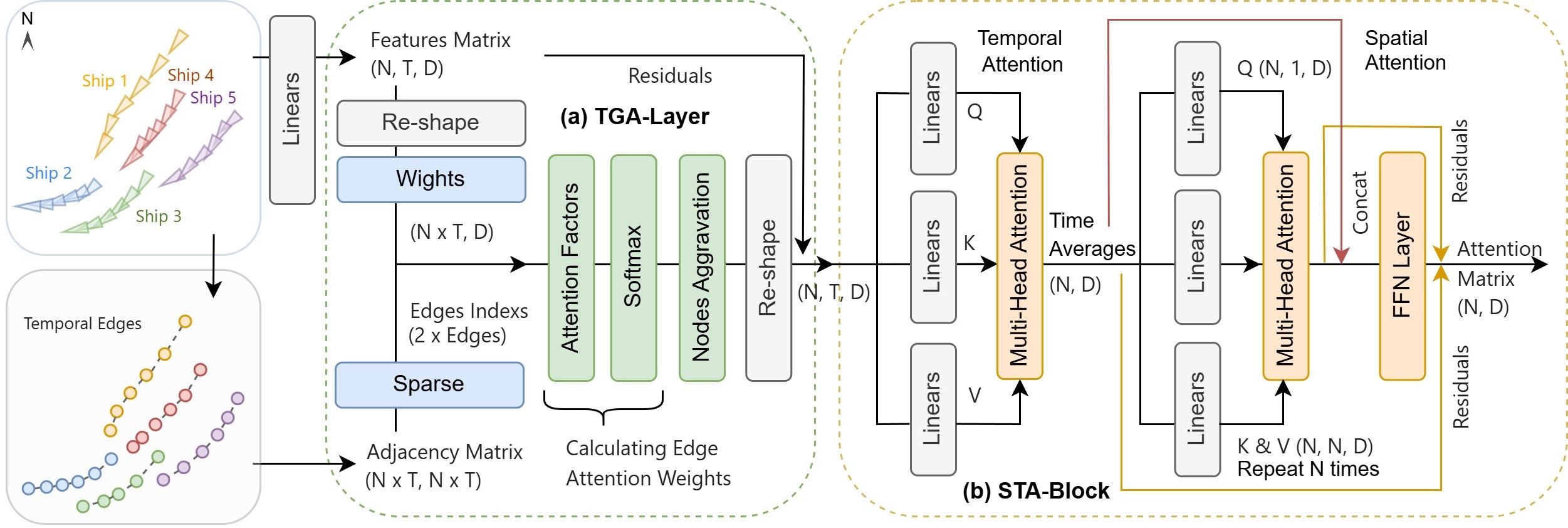}
        \caption{The pipeline of the proposed TGA layer and STA block. The TGA layer processes dynamic temporal graphs using a graph attention mechanism. The STA block performs three key operations: spatial feature extraction, temporal feature extraction, and feature fusion through a feed-forward network.}
        \label{figure:stab}
    \end{figure*}
\subsection{Graph Representation}
    To effectively capture the complex relationships between observations from different perspectives, we transform the trajectories of AIS and CCTV data into structured graph representations. As shown in Fig. \ref{figure:framework}, a dynamic graph $G = (V, E)$ is constructed for each observation set, where $V$ denotes the set of nodes and $E$ represents the set of edges. The graph encodes temporal evolution patterns, enabling the modeling of dynamic behaviors over time. Specifically, each node $v_i \in V$ corresponds to an individual observation, the detailed structure of which is provided in Eq. (\ref{eq:observation}). The edge set $E$ captures temporal relationships through two types of connections. Temporal edges encode the evolution of relationships across consecutive time steps. Formally, for nodes $V_{t-1}$ at time $t-1$ and $V_t$ at time $t$, the temporal edges $\mathcal{E}$ are defined as follows
    \begin{equation}
        \mathcal{E} = \{(v_i, v_j), (v_j, v_i) \mid v_i \in V_{t-1}, v_j \in V_t\},
    \end{equation}
    where $(v_i, v_j)$ represents a directed edge from node $v_i$ at time $t-1$ to node $v_j$ at time $t$, while $(v_j, v_i)$ denotes the reverse relationship. These bidirectional temporal edges effectively capture the dynamic interactions between observations across consecutive time steps, enabling prompt correction of error matches occurring at one time step in subsequent steps.
\section{Graph Learning-driven Multi-vessel Association}\label{sec:ourmethod}
\subsection{Overview}
    As shown in Fig. \ref{figure:framework}, our framework consists of four components proposed to progressively refine the matching process. First, the TGA layer constructs dynamic graphs that capture temporal relationships between vessels, representing multimodal trajectories. These graphs serve as inputs to a STA block, which learns discriminative feature representations. The extracted features are then fused to compute similarity scores for potential matches by MLP-UMF. Finally, the similarity scores are optimized using the Hungarian algorithm to ensure globally consistent assignments.
\subsection{Temporal Graph Attention Layer}
    To model the inherent sequential dependencies in vessel trajectories, we propose the TGA layer. By constructing directed temporal graphs where nodes represent vessel states and edges encode temporal relationships, the TGA layer can effectively capture both local motion dynamics and long-range patterns crucial for trajectory modeling. Specifically, as shown in the Fig. \ref{figure:stab} (a), the node features $h \in \mathbb{R}^{N_n \times T \times d}$, where $N_n$, $T$, and $d$ are the number of nodes, the number of time steps, and the feature dimension, respectively, are then updated through multi-head attention, which can be given as
    \begin{equation}
        h_i^{(l)} = \sigma_{\text{ReLU}}(\frac{1}{K}\sum_{k=1}^K \sum_{j \in \mathcal{T}(i)} \beta_{ij}^k W^k h_j^{(l-1)}),
    \end{equation}
    where $h_i^{(l)}$ and $h_j^{(l-1)}$ denote the feature vectors of the central node $i$ and its temporal neighbor $j$, respectively, at the $l$-th layer. $\sigma_{\text{ReLU}}$ is the rectified linear unit, $K$ is the number of attention heads, $\mathcal{T}(i)$ denotes the temporal neighbors of node $i$, and $W^k$ is the learnable projection matrix for the $k$-th attention head. The attention weights $\beta_{ij}^k$ adaptively emphasize relevant temporal contexts, which can be given as
    \begin{equation}\label{eq:beta}
        \beta_{ij}^k = \text{softmax}(\frac{(W_Q^k h_i^{(l-1)})^T(W_K^k h_j^{(l-1)})}{\sqrt{d_k}}),
    \end{equation}
    where $W_Q^k$ and $W_K^k$ are learnable projection matrices for queries and keys, respectively, and $d_k = d/K$ is the dimensionality of each attention head. The TGA layer dynamically aggregates information from spatial and temporal neighbors, allowing the model to encode hierarchical motion features. It can effectively extract both local motion dynamics (e.g., changes in speed or direction) and long-range dependencies (e.g., route trends across time).
\subsection{Spatial-Temporal Attention Block}
    The STA block is proposed to learn discriminative features that extract the spatial and temporal dynamics of vessel motion from AIS and CCTV. Within the GMvA framework, dynamic graphs are processed through the STA block (shown in the Fig. \ref{figure:stab} (b)), which performs three key operations: spatial feature extraction, temporal feature extraction, and feed-forward network (FFN)-based spatiotemporal feature fusion.
    \begin{figure*}[t]
        \centering
        \setlength{\abovecaptionskip}{0.cm}
        \includegraphics[width=0.975\linewidth]{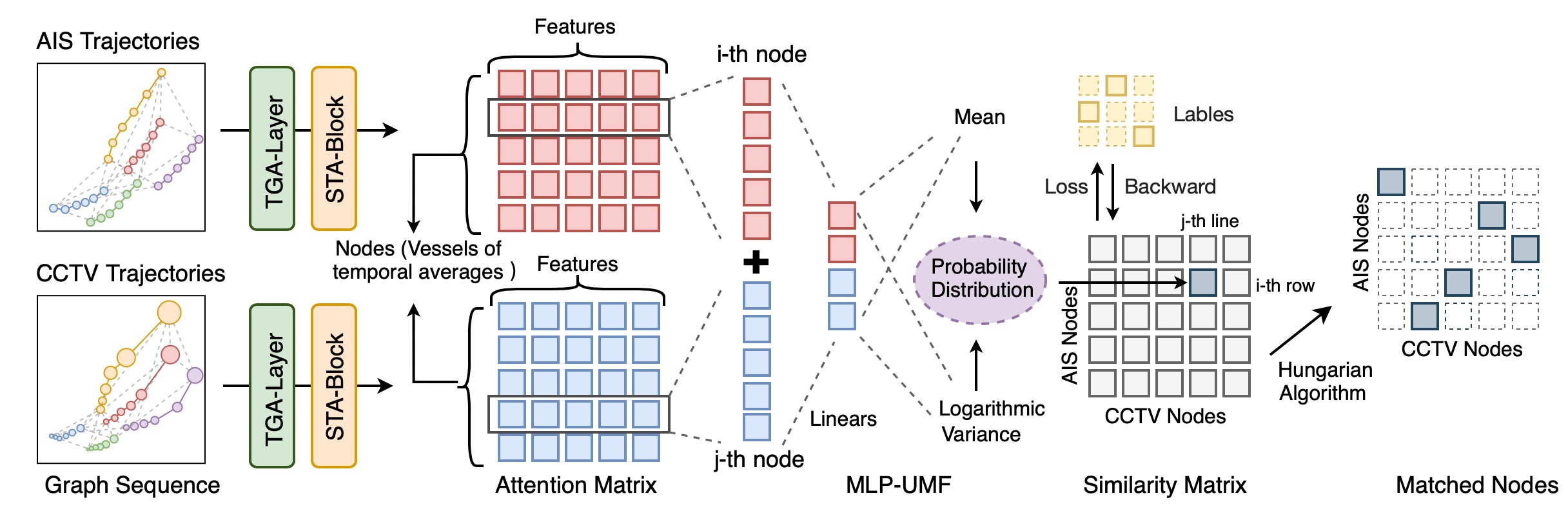}
        \caption{The pipeline of feature fusion and similarity computation. The STA block-generated features are then fused through a dedicated MLP-UMF, which computes similarity scores between potential matches by seamlessly incorporating both appearance and geometric constraints.}
        \label{figure:ff}
    \end{figure*}
\subsubsection{Temporal Feature Extraction}
    To extract the dynamic temporal dependencies $h_i^t$ inherent in vessel motion, we suggest a multi-head temporal attention mechanism ($\text{MultiHead}(\cdot)$), which is defined as
    \begin{equation}
        h_i^t = \text{MultiHead}(h_i^{(l)}, \{h_j^{(l)}\}_{j \in \mathcal{T}(i)}),
    \end{equation}
    where $\mathcal{T}(i)$ denotes the temporal neighborhood of node $i$. This mechanism models vessel-specific temporal behaviors, such as gradual acceleration, deceleration, and heading adjustments, or sudden changes in motion due to environmental factors (e.g., obstacles). Each attention head computes temporal relationships independently, and the $k$-th head is
    \begin{equation}
        \text{head}_k = \sum_{j \in \mathcal{T}(i)} \beta_{ij}^k W_t^k h_j^{l},
    \end{equation}
    where $\beta_{ij}^k$ are the temporal attention weights (as detailed in Eq. (\ref{eq:beta})), which assign dynamic importance to past observations based on their relevance to the current state of the vessel. 
    
    The temporal feature extraction allows the model to prioritize critical temporal transitions. By assigning dynamic importance to past states, the multi-head temporal attention effectively captures both short-term variations (e.g., sudden evasive actions) and long-term dependencies (e.g., gradual heading adjustments due to inertia).
\subsubsection{Spatial Feature Extraction}
    The spatial feature extraction module is proposed to aggregate information from spatially adjacent nodes while retaining the unique features of the central node. It can reflect key vessel behaviors such as proximity-based interactions, collision avoidance, and coordinated movements within fleets. The aggregated spatial representation $h_i^s$ is given as
    \begin{equation}
        h_i^s = \text{LN}(\sum_{j \in \mathcal{N}(i)} \alpha_{ij} W_s^{(l)} h_j^{(l-1)} + h_i^{(l-1)}),
    \end{equation}
    where $\text{LN}(\cdot)$ represents layer normalization, $\mathcal{N}(i)$ denotes the set of spatial neighbors of node $i$, $W_s^{(l)}$ is a learnable spatial transformation matrix, and $\alpha_{ij}$ is the attention coefficient quantifying the importance of neighbor $j$ relative to node $i$. The attention mechanism enables the model to adaptively prioritize interactions based on vessel-specific dynamics, such as vessels maintaining safe distances or forming traffic patterns in narrow waterways. The attention coefficient $\alpha_{ij}$ is given as
    \begin{equation}
        \alpha_{ij} = \frac{\exp\left(g(W_q h_i^{(l-1)}, W_k h_j^{(l-1)})\right)}{\sum_{k \in \mathcal{N}(i)} \exp\left(g(W_q h_i^{(l-1)}, W_k h_j^{(l-1)})\right)},
    \end{equation}
    where $g(\cdot, \cdot)$ is a learnable scoring function that evaluates the compatibility between nodes, and $W_q, W_k$ are learnable query and key matrices. The scoring function captures vessel-specific spatial interactions, such as the relative heading alignment between vessels, their distance, and velocity vectors.
\subsubsection{FFN-based Spatiotemporal Feature Fusion}
    The final stage of STA block integrates spatial and temporal features to capture the complex interplay between vessel interactions and motion dynamics. The fused representation is computed as
    \begin{equation}
        h_i^{f} = \text{FFN}(\text{LN}([h_i^t || h_i^s])) + h_i^t + h_i^s,
    \end{equation}
    where $[h_i^t || h_i^s]$ denotes the concatenation of temporal features $h_i^t$ and spatial features $h_i^s$. This fusion reflects the integration of localized vessel interactions (e.g., proximity, heading alignment) with temporal patterns (e.g., velocity changes, course adjustments). The FFN operation is defined as
    \begin{equation}
        \text{FFN}(x) = W_2(\sigma_{\text{ReLU}}(W_1x + b_1)) + b_2,
    \end{equation}
    where $W_1$ and $W_2$ are learnable weight matrices, and $b_1$ and $b_2$ are biases. The FFN introduces non-linear transformations, enabling the model to learn complex relationships between spatial and temporal features.  
\subsection{MLP-based Uncertainty Fusion Module}
    \begin{algorithm}[t]
    \caption{Similarity and Matching Optimization}
    \label{alg:matching}
    \begin{algorithmic}[1]
    \Require Feature matrices $H_A$, $H_B$
    \Ensure Matching matrix $\mathcal{M}$
    \Function{Compute Matching Matrix}{$H_A$, $H_B$}
        \State // Compute uncertainty-aware similarity matrix
        \For{$i \gets 1$ to $N_A$}
            \For{$j \gets 1$ to $N_B$}
                \State $f_{ij} \gets$ [Cat($h_i^A$, $h_j^B$, CGF($i$, $j$))]
                \State $S_{ij} \gets$ MLP-UFM($f_{ij}$)
            \EndFor
        \EndFor
        \State // Apply matching threshold
        \ForAll{$(i,j)$ where $S_{ij} < \tau$}
            \State $S_{ij} \gets -\infty$
        \EndFor
        \State // Hungarian algorithm optimization
        \State $\mathcal{M} \gets$ Hungarian Algorithm
        \State \Return $\mathcal{M}$
    \EndFunction
    \end{algorithmic}
    \end{algorithm}
    To compute multiple similarity scores between potential matches, as shown in Fig. \ref{figure:ff}, we propose a MLP-based uncertainty fusion module (MLP-UFM), as described in Algorithm \ref{alg:matching}. For each candidate pair $(i, j)$, a joint feature vector $f_{ij}$ is constructed to encode various aspects of similarity relevant to vessel interactions and behaviors
    \begin{equation}
        f_{ij} = [h_i^{f_A} || h_j^{f_B} || d_{ij} || \phi_{ij}],
    \end{equation}
    where $h_i^{f_A}$ and $h_j^{f_B}$ are the learned node representations from graphs $A$ and $B$, respectively, encoding the unique characteristics of vessels or trajectory segments. The term $d_{ij}$ represents spatial proximity, modeling the relative distance between vessels, while $\phi_{ij}$ captures additional contextual or behavioral features.
    
    To enhance robustness, we integrate uncertainty into the similarity computation by modeling the similarity score for each pair $(i, j)$ as a Gaussian distribution $\mathcal{N}(\mu_{ij}, \sigma_{ij}^2)$. The mean $\mu_{ij}$ and the log variance $\text{logvar}_{ij}$ are predicted independently using two separate multi-layer perceptrons (MLPs), respectively. Specifically, the mean $\mu_{ij}$ represents the predicted central similarity score, while the variance $\sigma_{ij}^2$ (derived from the log variance $\text{logvar}_{ij}$) captures the associated uncertainty. To ensure numerical stability, the variance is computed as
    \begin{equation}
        \sigma_{ij}^2 = \exp(\text{logvar}_{ij}).
    \end{equation}
    
    The uncertainty-aware similarity score for each pair is computed by normalizing the predicted mean similarity scores with their corresponding variance, thereby penalizing predictions with higher uncertainty $S_{ij}$,
    \begin{equation}    
        S_{ij} = \operatorname{Sigmoid}(\frac{\mu_{ij}}{\sigma_{ij}^2})
    \end{equation}
        
    The uncertainty-aware framework balances similarity predictions with associated uncertainty, resulting in a robust and interpretable similarity matrix. The normalized similarity score $S_{ij}$ inherently includes the effect of the sigmoid activation function, ensuring it falls within the range $[0, 1]$.
    \begin{figure*}[t]
        \centering
        \setlength{\abovecaptionskip}{0.cm}
        \includegraphics[width=0.975\linewidth]{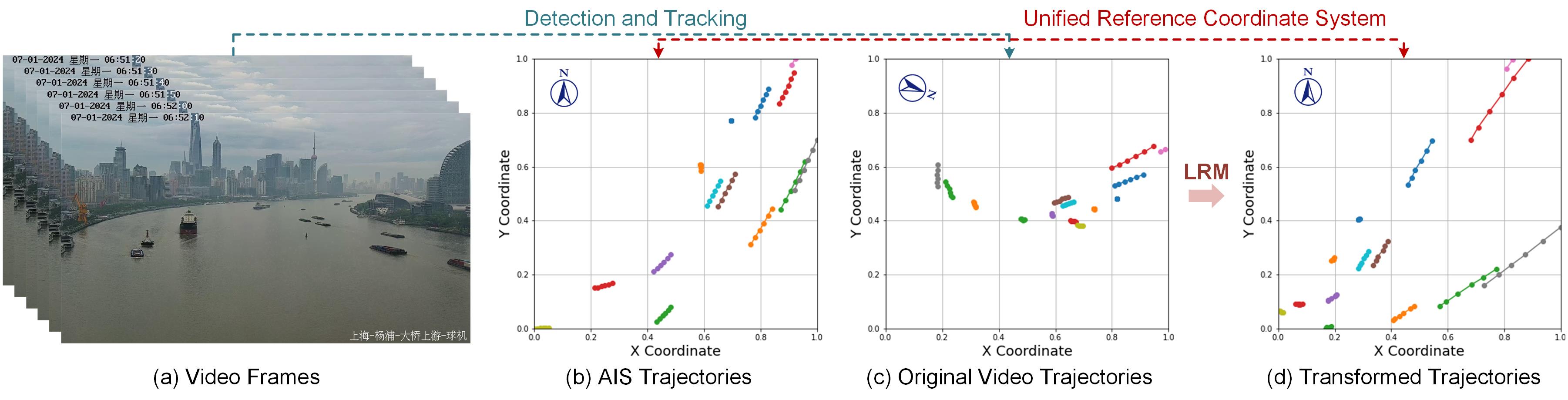}
        \caption{To unify the coordinate systems of AIS and video trajectories, we use a linear regression model (LRM) for coordinate transformation. The unified coordinate systems (i.e., (b) and (d)) have similar trajectory spatial distributions. GMvA focuses more on the relative spatiotemporal features of multiple targets to reduce the accuracy loss caused by coordinate conversion errors.}
        \label{figure:c}
    \end{figure*}
\subsection{Matching Optimization}
    In the final stage, our model transforms similarity scores into optimal matching assignments, ensuring global consistency throughout the process. The matching problem is formulated as a maximum weighted bipartite matching
    \begin{equation}
        \mathcal{M}^* = \arg\max_{\mathcal{M}} \sum_{i,j} S_{ij}\mathcal{M}_{ij},
    \end{equation}
    subject to one-to-one matching constraints
    \begin{equation}
        \sum_i \mathcal{M}_{ij} \leq 1, \sum_j \mathcal{M}_{ij} \leq 1, \mathcal{M}_{ij} \in \{0,1\}.
    \end{equation}

    We solve this optimization problem efficiently using the Hungarian algorithm, augmented with additional constraints to handle outliers and unmatched targets
    \begin{equation}
        \mathcal{M}_{ij} = 0 \text{ if } S_{ij} < \tau,
    \end{equation}
    where $\tau$ is a learned threshold to adapt different scenarios.
    %
    %
%
\subsection{Multi-component Loss Function}
    Our model uses a multi-component loss function to tackle various aspects of the matching problem. The total loss $\mathcal{L}_{\text{total}}$ is composed of two primary components: matching loss $\mathcal{L}_{\text{match}}$, and contrastive loss $\mathcal{L}_{\text{con}}$, i.e.,
    \begin{equation}
        \mathcal{L}_{\text{total}}(S_{ij},T_{ij}) = \gamma_1\mathcal{L}_{\text{match}}(S_{ij},T_{ij}) + \gamma_2\mathcal{L}_{\text{con}}(S_{ij}),
    \end{equation}
    where $\gamma_1=1.0$ and $\gamma_2=0.5$ are weighting coefficients that balance the contributions of the match and contrastive terms.
\subsubsection{Matching Loss}
    The matching loss $\mathcal{L}_{\text{match}}$ serves as the primary optimization objective. It adopts a cross-entropy formulation to measure the discrepancy between predicted and truth matching scores. For each pair of samples $(i,j)$, we compute the log loss between the predicted probability $S_{ij}$ and the true label $T_{ij}$, averaged over all $N$ sample pairs
    
    \begin{equation}
        \mathcal{L}_{\text{match}} = -\frac{1}{N}\sum_{i,j} (T_{ij}\log(S_{ij}) + (1-T_{ij})\log(1-S_{ij})).
    \end{equation}
\subsubsection{Contrastive Loss}
    To enhance the model's discriminative power, we incorporate a contrastive loss $\mathcal{L}_{\text{con}}$. It is suggested to ensure that similarity scores between positive pairs significantly exceed those of negative pairs. Specifically, for each positive pair $(i,j)$, we require its similarity score $S_{ij}$ to be higher than the corresponding negative error pair score $S_{ie}$ by a margin $m$, i.e.,
    \begin{equation}
        \mathcal{L}_{\text{con}} = \frac{1}{N_p}\sum_{(i,j)\in \mathcal{P}} \max(0, m + S_{ij} - S_{ie}),
    \end{equation}
    where $\mathcal{P}$ represents the set of all positive pairs, $N_p$ is the number of positive pairs, and $m$ is a preset margin parameter controlling the minimum gap between positive and negative pair scores.
\section{Experiments and Discussion}\label{sec:experiments}
    This section details the experimental setup and results. We describe the training/testing datasets, coordinate system, experimental platform, competitive methods, and evaluation metrics. We then analyze GMvA's performance, comparing it with competitive methods using quantitative and qualitative results. Ablation studies validate the network's structure and key components. Finally, we assess GMvA's running time and computational complexity under three vessel density scenarios.
\subsection{Implementation Details}
\subsubsection{Dataset}
    This work selected different waterway sections of the Huangpu River in Shanghai, China as the research area. During the data preprocessing phase, we first performed regional filtering and data cleaning on the AIS data, and unified the sampling frequency to once every 10 seconds through cubic spline interpolation. For video processing, we utilized the framework described in Literature \cite{guo2023asynchronous} to detect vessels and assign each one a unique tracking identifier (ID). Subsequently, we established the correspondence between AIS data and video detection results through manual annotation, providing a standard dataset for model training and testing. Temporal consistency was ensured by synchronizing the sampling times of both AIS and video data to multiples of 10 seconds, yielding six sampling points per minute. The data was then sorted by timestamp and target IDs, grouped, and prepared for further analysis. A sliding window method is applied to divide the time sequences into fixed-length time windows (default size is 6). Each window contains target data within the corresponding time range for graph feature construction. 
    \begin{figure}[t]
        \centering
        \setlength{\abovecaptionskip}{0.cm}
        \includegraphics[width=0.900\linewidth]{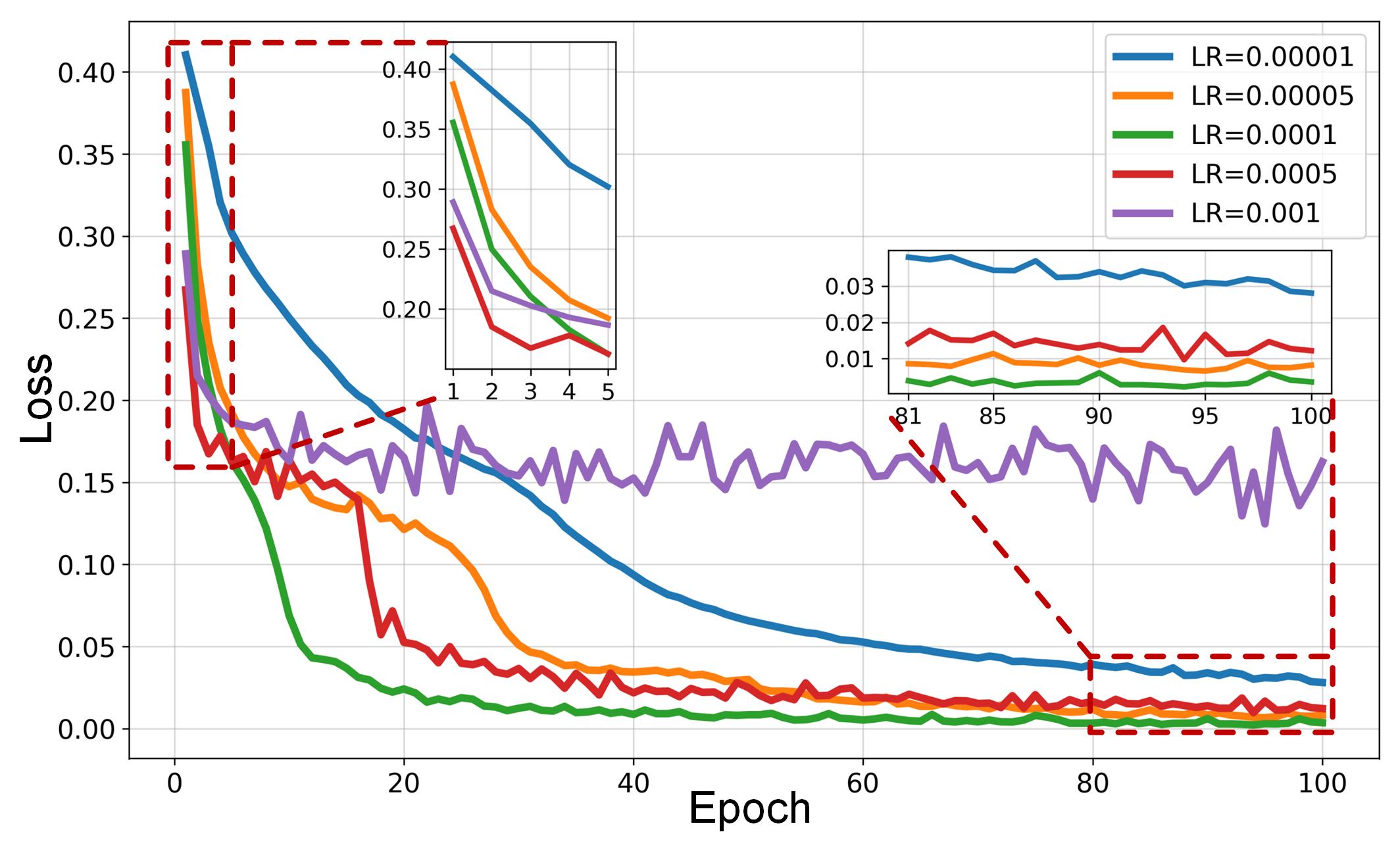}
        \caption{The loss convergence of five different learning rates (i.e., 0.00001, 0.00005, 0.0001, 0.0005, 0.001) over 100 epochs. The main plot provides an overview of the entire training process, while the inset focuses on the first 5 and last 20 epochs.}
        \label{figure:LR}
    \end{figure}
\subsubsection{Unified Coordinate System}\label{subsubsection:ucs}
    To achieve a unified coordinate system for AIS and video trajectories, a linear regression model (LRM) is utilized for coordinate transformation. Its process consists of two primary steps. First, a mapping relationship between the two coordinate systems is established based on matched AIS and video data samples collected across multiple time points. It also be aligned by finding reference objects with known specific positions in the image. Subsequently, this mapping is applied to transform all video coordinates into the AIS coordinate system. The transformed coordinates are then employed to generate time-series trajectory plots with consistent spatial references. The transformation process is expressed mathematically as
    \begin{equation}
        \begin{bmatrix} 
        x_i^b \\ 
        y_i^b 
        \end{bmatrix}
        =
        \begin{bmatrix} 
        w_{x1} & w_{x2} \\ 
        w_{y1} & w_{y2} 
        \end{bmatrix}
        \cdot
        \begin{bmatrix} 
        \hat{x}_i^b \\ 
        \hat{y}_i^b
        \end{bmatrix}
        +
        \begin{bmatrix} 
        b_x \\ 
        b_y 
        \end{bmatrix},
    \end{equation}
    where $\hat{x}_i^b$ and $\hat{y}_i^b$ denote the pixel coordinates of the target in the video frame, $w_{x1}$, $w_{x2}$, $w_{y1}$, and $w_{y2}$ represent the weights that quantify the influence of the input features on the target values, and $b_x$ and $b_y$ are the biases. Fig. \ref{figure:c} shows the result after unifying the coordinates. While minor accuracy loss may occur during the coordinate transformation process, it eliminates the dependency on specific camera and position parameters for coordinate conversion. Moreover, as the GMvA emphasizes the relative positional relationships between multiple targets rather than the absolute precision of the transformed coordinates, the impact of such accuracy loss is further mitigated.
    \begin{figure}[t]
        \centering
        \setlength{\abovecaptionskip}{0.cm}
        \includegraphics[width=0.900\linewidth]{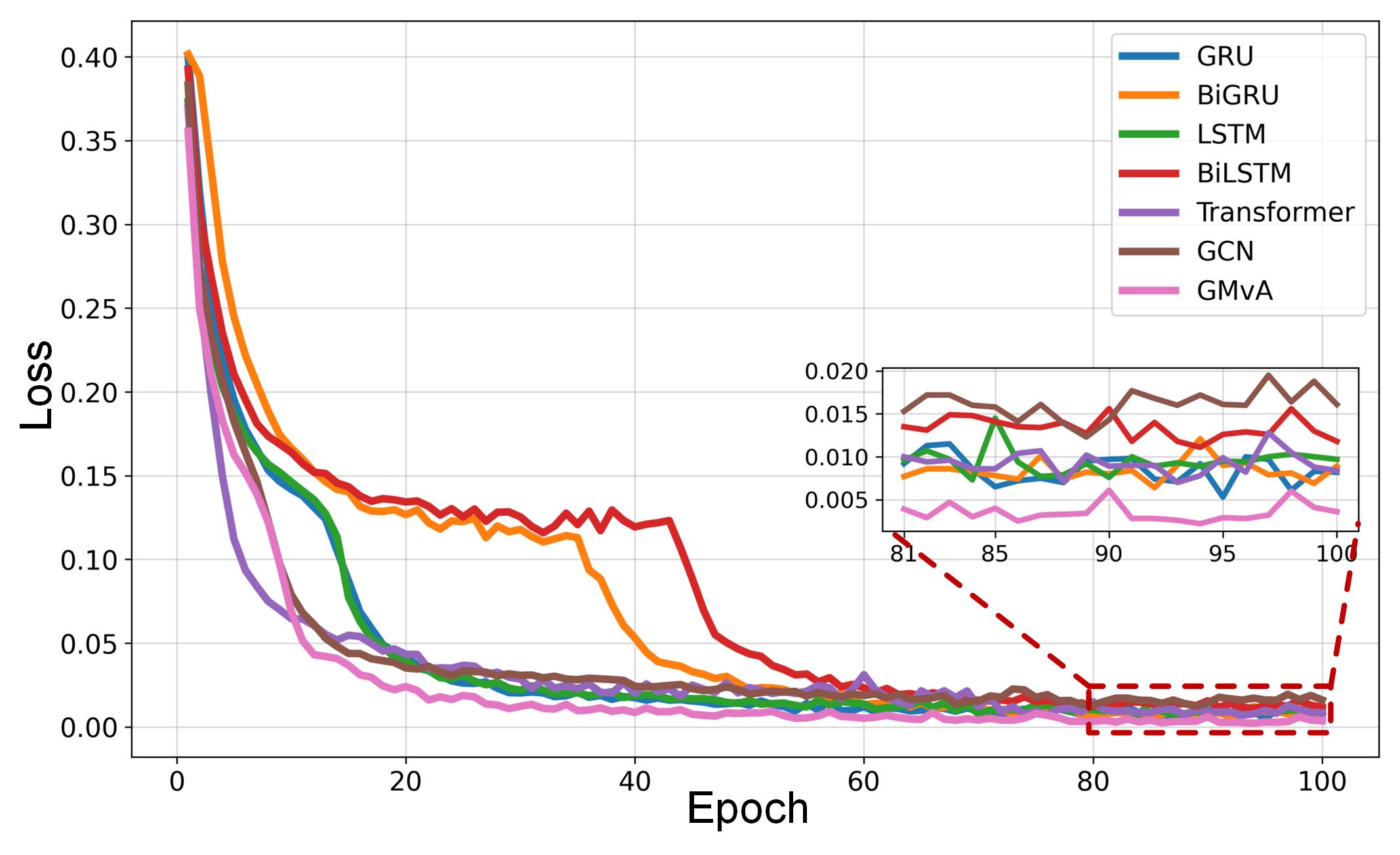}
        \caption{The loss convergence of seven different models (i.e., GRU, BiGRU, LSTM, BiLSTM, Transformer, GCN, and our GMvA) over 100 epochs with a learning rate of 0.0001. The main plot provides an overview of the entire training process, while the inset focuses on the last 20 epochs.}
        \label{figure:Convergence}
    \end{figure}
    \setlength{\tabcolsep}{8.00pt}
    \begin{table}[t]
        \centering
        \footnotesize
        \caption{Performance evaluation of the GMvA and other competitive methods under different vessel density scenarios. The best results are in \textbf{bold}, and the second best are with \underline{underline}.}
        \begin{tabular}{l|ccc|c}
        \hline
        Methods     & \makecell {Low- \\ Density} & \makecell {Moderate- \\ Density}   & \makecell {High- \\ Density}   & Average   \\\hline\hline
            ED          & 88.07\% & 75.45\% & 57.32\% & 73.61\% \\
            CD          & 76.71\% & 56.77\% & 40.13\% & 57.87\% \\
            PDF         & 89.67\% & \underline{81.49\%} & \underline{77.53\%} & 82.90\% \\
            DTW         & 56.83\% & 39.06\% & 27.64\% & 41.18\% \\
            GRU \cite{chung2014empirical}         & 62.99\% & 48.26\% & 30.04\% & 47.10\% \\
            BiGRU \cite{cho2014learning}       & 62.82\% & 43.69\% & 26.52\% & 44.34\% \\
            LSTM \cite{hochreiter1997long}        & 80.33\% & 74.80\% & 71.80\% & 75.64\% \\
            BiLSTM \cite{schuster1997bidirectional}      & 85.88\% & 63.23\% & 52.07\% & 67.06\% \\
            Transformer \cite{giuliari2021transformer} & \underline{97.02\%} & 77.45\% & 74.24\% & \underline{82.90\%} \\
            GCN \cite{kipf2016semi}         & 90.56\% & 75.57\% & 64.72\% & 76.95\% \\\hline
            GMvA        & \textbf{98.51\%} & \textbf{91.61\%} & \textbf{86.73\%} & \textbf{92.28\%}\\\hline
        \end{tabular}\label{table:dres1}
    \end{table}
    \setlength{\tabcolsep}{9.50pt}
    \begin{table*}[t]
        \centering
        \footnotesize
        \caption{Performance evaluation of the GMvA and other competitive methods under different missing vessel number scenarios. The best results are in \textbf{bold}, and the second best are with \underline{underline}.}
        \begin{tabular}{l|cccccc|c}
        \hline
        Methods     & No Missing  & Two Missing   & Four Missing  & Six Missing   & Eight Missing   & Ten Missing   & Average \\\hline\hline
        ED          & 57.32\% & 44.03\% & 38.40\% & 32.71\% & 29.42\% & 26.32\% & 38.03\% \\
        CD          & 40.13\% & 33.25\% & 30.00\% & 26.57\% & 27.59\% & 23.05\% & 30.10\% \\
        PDF         & \underline{77.53\%} & 61.05\% & \underline{57.93\%} & 50.49\% & 37.88\% & 28.58\% & 52.24\% \\
        DTW         & 27.64\% & 23.92\% & 22.20\% & 20.47\% & 18.97\% & 16.12\% & 21.55\% \\
        GRU \cite{chung2014empirical}         & 30.04\% & 18.41\% & 21.84\% & 23.03\% & 23.62\% & 26.53\% & 23.91\% \\
        BiGRU \cite{cho2014learning}       & 26.52\% & 22.00\% & 27.33\% & 28.27\% & 30.37\% & 32.31\% & 27.80\% \\
        LSTM \cite{hochreiter1997long}        & 71.80\% & 58.37\% & 54.10\% & \underline{51.40\%} & \underline{43.63\%} & 38.38\% & \underline{52.95\%} \\
        BiLSTM \cite{schuster1997bidirectional}      & 52.07\% & 23.95\% & 21.15\% & 20.14\% & 19.17\% & 23.98\% & 26.74\% \\
        Transformer \cite{giuliari2021transformer} & 74.24\% & \underline{66.66\%} & 51.15\% & 44.12\% & 32.73\% & 18.26\% & 47.86\% \\
        GCN \cite{kipf2016semi}         & 64.72\% & 34.67\% & 37.27\% & 35.76\% & 39.17\% & \underline{41.68\%} & 42.21\% \\\hline
        GMvA        & \textbf{86.73\%} & \textbf{75.31\%} & \textbf{60.95\%} & \textbf{56.95\%} & \textbf{51.67\%} & \textbf{45.45\%} & \textbf{62.84\%} \\\hline
                \end{tabular}\label{table:dres}
    \end{table*}
\subsubsection{Optimizer and Experimental Setup}
    The Adam optimizer is used during training, with a learning rate of $0.0001$ to ensure stable and efficient parameter updates. The model is trained for 100 epochs, with parameters optimized iteratively across time windows. To mitigate potential issues with gradient explosion, gradient clipping is applied to constrain the maximum norm of the gradients. As shown in Fig. \ref{figure:LR}, the convergence of our method under different learning rates demonstrates that a learning rate of 0.0001 yields the best performance. All experiments are conducted in a Python 3.9 environment using the PyTorch software library. The computational setup includes a PC equipped with an Intel(R) Core(TM) i9-12900K CPU @ 2.30GHz and an Nvidia GeForce RTX 4090 GPU, ensuring efficient execution of training and inference tasks.
    \begin{figure*}[t]
        \centering
        \setlength{\abovecaptionskip}{0.cm}
        \includegraphics[width=0.975\linewidth]{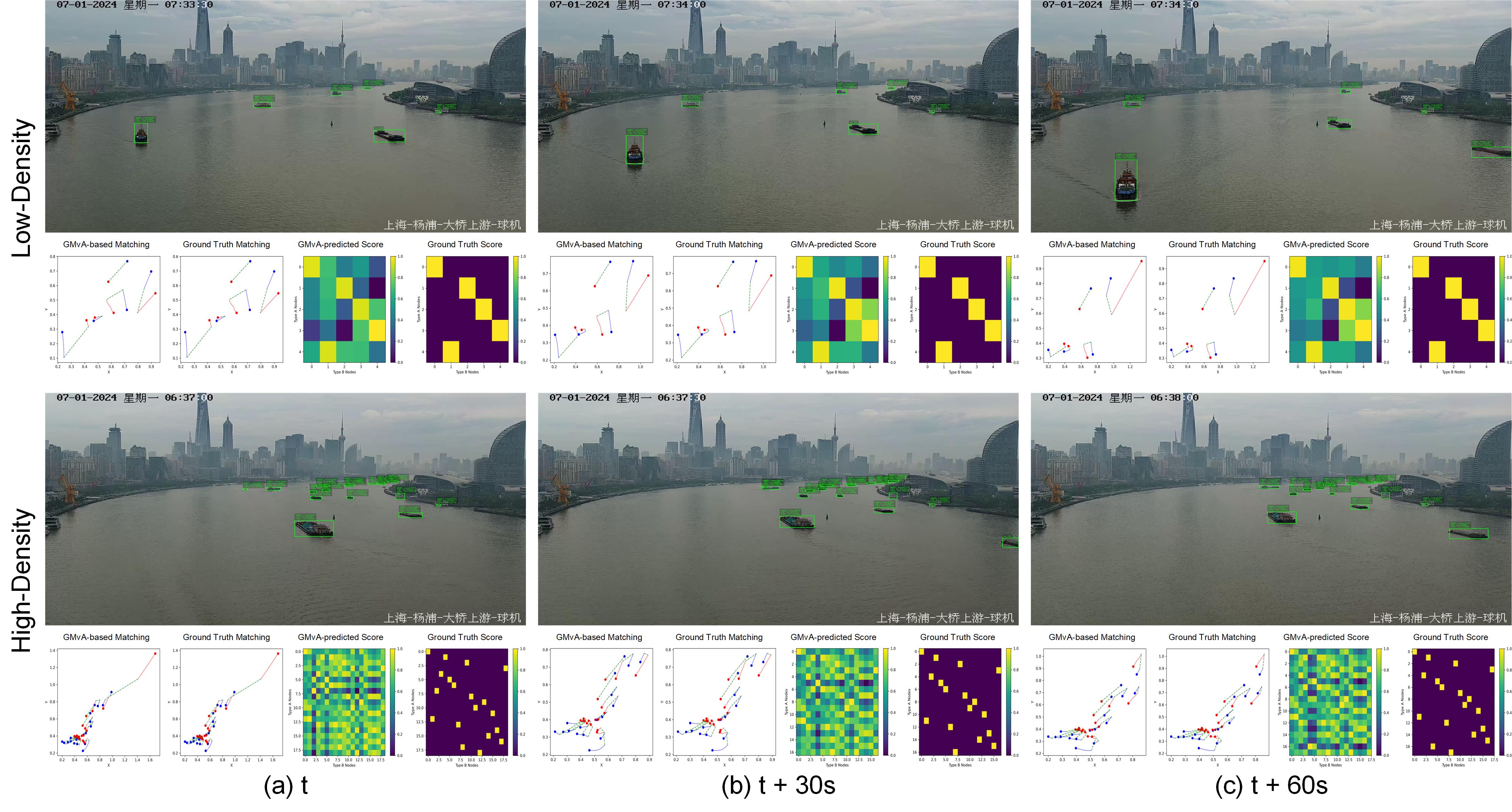}
        \caption{Visualization of multi-vessel association under low and high vessel density. In high-density scenarios, as vessels move farther from the camera, their apparent size diminishes, and a clustering phenomenon becomes apparent near curved sections of the waterways, posing a greater challenge to target association. High-resolution viewing is recommended for detailed inspection.}
        \label{figure:QA}
    \end{figure*}
\subsubsection{Competitive Methods}
    To demonstrate the performance of our proposed method, we selected a variety of methods from the table for comparison, including Euclidean Distance (ED), Chebyshev Distance (CD), Dynamic Time Warping (DTW), Probability Density Function (PDF), as well as several deep learning models such as Gated Recurrent Unit (GRU) \cite{chung2014empirical}, Bidirectional Gated Recurrent Unit (BiGRU) \cite{cho2014learning}, Long Short-Term Memory (LSTM) \cite{hochreiter1997long}, Bidirectional Long Short-Term Memory (BiLSTM) \cite{schuster1997bidirectional}, Transformer \cite{giuliari2021transformer}, and Graph Convolutional Network (GCN) \cite{kipf2016semi}. To ensure the objectivity and fairness of the experiments, all methods were implemented using the same dataset and identical training parameters, with the test dataset strictly excluded from the training process.
\subsubsection{Performance Evaluation}
    In the maritime target matching task with multimodal heterogeneous data fusion, accuracy $\text{Acc}$ evaluates the performance of the cross-modal target association method as follows
    \begin{equation}
        \text{Acc} = \frac{|M_{pred} \cap M_{true}|}{|M_{true}|} \times 100 \%,
    \end{equation}
    where $M_{pred}$ and $M_{true}$ denote the set of predicted and real matches between AIS and video targets, $|\cdot|$ represents the cardinality of a set, and $\cap$ represents set intersection.

    In the qualitative evaluation, we will incorporate AIS information into the images to visually present the comparison between the GMvA-matched and the ground truth matched trajectory points, as well as the corresponding heatmap of the matching matrix. To ensure consistency in the analysis, both the matched and the ground truth matched trajectory points will be uniformly represented using the true north direction.
\subsection{Quantitative Analysis}
\subsubsection{Convergence Analysis}
    Fig. \ref{figure:Convergence} illustrates the convergence of training loss for seven different learning models, over the course of 100 training epochs. The main plot provides an overview of the entire training process, clearly showing the general trend and rate of convergence for each model. It highlights the significant differences in loss reduction across the models during the initial epochs, where the majority of loss decay occurs. To facilitate a more granular comparison of the models' fine-tuning behavior near convergence, an inset is included that zooms in on the loss trajectories during the last 20 epochs. This detailed view enables the identification of subtler differences in the convergence rates and final loss values achieved by the models. Notably, the Transformer model demonstrates the most rapid and consistent convergence, achieving the lowest loss among all models by the end of training. In contrast, GMvA also exhibits competitive performance with a steady decline in loss, particularly during later epochs, suggesting its potential for robust optimization.
    \begin{figure*}[t]
        \centering
        \setlength{\abovecaptionskip}{0.cm}
        \includegraphics[width=0.975\linewidth]{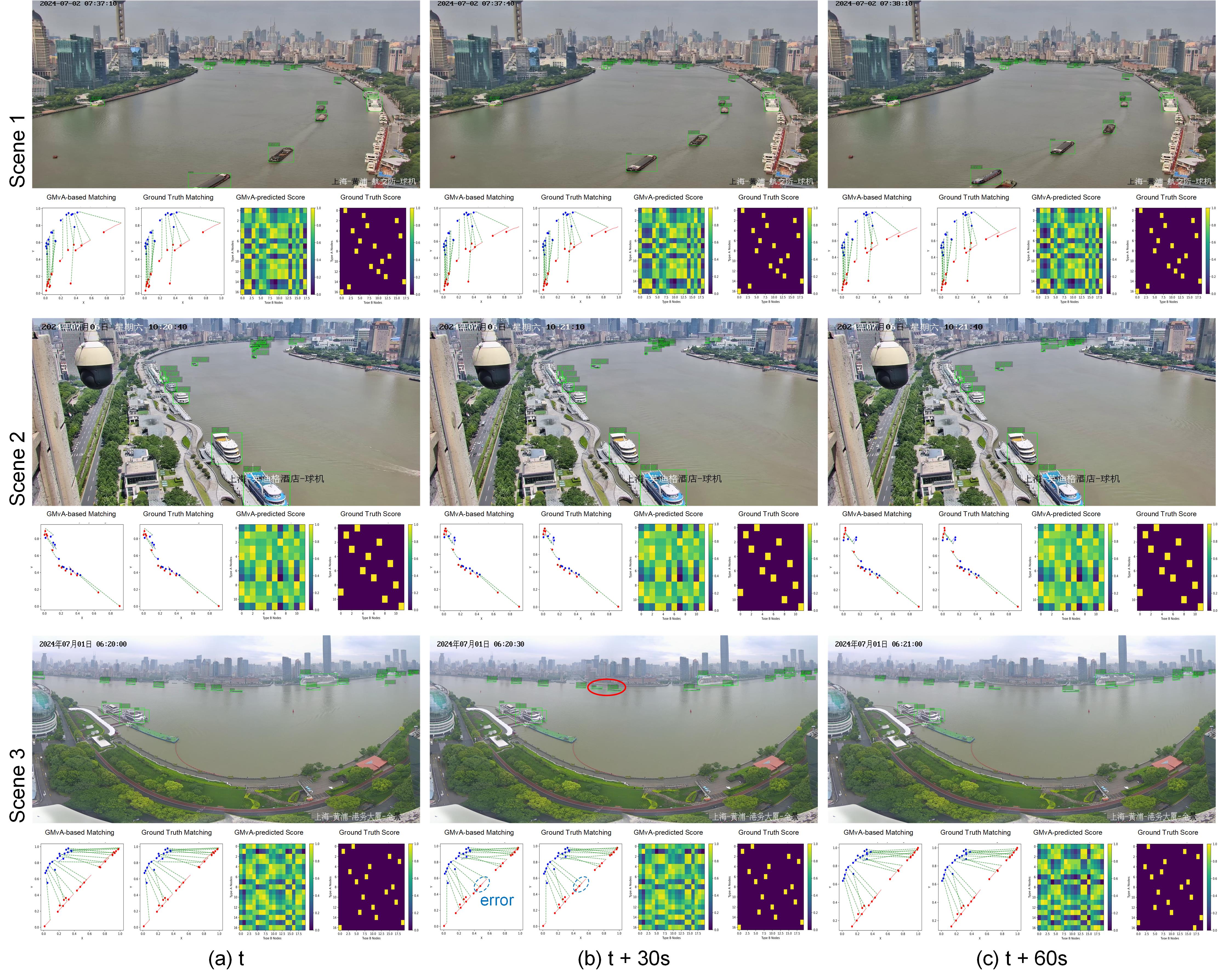}
        \caption{Visualization of multi-vessel trajectories association under different scenes. Our GMvA has stable target association performance across various scenarios when fewer vessels are present. Even if a matching error occurs in Area 3 (b), GMvA can be swiftly corrected at a later stage. High-resolution viewing is recommended for detailed inspection.}
        \label{figure:3s}
    \end{figure*}
\subsubsection{Impact of Vessel Density}
    To evaluate the performance of the model under different vessel density environments, we conducted multi-scenario experiments, including low density (approximately 10 vessels), moderate density (approximately 20 vessels), and high density (approximately 30 vessels). As shown in Table \ref{table:dres1}, in low-density scenarios, the matching accuracy of the model is close to perfect due to the large spatial intervals between targets. It is primarily because the GNN can effectively capture the local relationship between targets. In moderate-density scenarios, despite the increase in the number of targets, the construction of the dynamic graph retains the temporal continuity and spatial correlation of the targets, allowing the model to stably output high matching accuracy. In high-density scenarios, traditional methods perform poorly due to target overlap or trajectory intersection. In contrast, our method models the complex relationship between targets through graph convolution and a temporal attention mechanism, significantly improving the matching accuracy.
\subsubsection{Impact of Data Missingness}
    In practical applications, missing data is a common challenge, such as when AIS signals are interrupted due to equipment failure or when video data is lost due to occlusion or poor quality. To test the robustness of the model under such conditions, we randomly deleted a specific number of AIS or video target records and evaluated the performance. As shown in Table \ref{table:dres}, it indicates that the matching performance of traditional methods dropped rapidly as the number of missing records increased. For instance, methods like ED and CD dropped to 26.32\% and 23.05\%, respectively, with ten missing records. In contrast, our proposed GMvA method maintained high accuracy, achieving 86.73\% with no missing data and retaining a significant 45.45\% accuracy even with ten missing records. This robustness is attributed to the ability of graph neural networks to fill in missing information through neighbor relationships and global features, while the temporal attention mechanism effectively captures trend changes in temporal data and avoids association failures caused by the loss of a single feature. These mechanisms enable GMvA to consistently outperform other competitive methods across all levels of missing data.
    \setlength{\tabcolsep}{8.00pt}
    \begin{table*}[t]
        \centering
        \footnotesize
        \caption{Results of the ablation study on module combinations.}
        \begin{tabular}{cc|ccccccc}\hline
            TGA Layer & STA Block & No Missing  & Two Missing   & Four Missing  & Six Missing   & Eight Missing & Ten Missing   & Average  \\\hline\hline
            \checkmark    &   & 55.55\% & 32.70\% & 32.71\% & 33.31\% & 36.27\% & 35.48\% & 37.67\% \\
            & \checkmark & 84.31\% & 67.11\% & 57.70\% & 52.94\% & 40.90\% & 33.29\% & 56.04\% \\\hline
            \checkmark    & \checkmark    & 86.73\% & 75.31\% & 60.95\% & 56.95\% & 51.67\% & 45.45\% & 62.84\% \\\hline
        \end{tabular}\label{table:as}
    \end{table*}
\subsection{Qualitative Analysis}
    Our method accurately matches AIS and video targets in two typical scenarios: sparse and busy waterways. As shown in Figs. \ref{figure:QA} and \ref{figure:3s}, the model effectively achieves target associations across scenarios of varying density and complexity. In low-density scenarios, where targets are fewer and more dispersed, the matching results completely align with the ground truth without errors or omissions. It demonstrates that the model is robust and reliable in handling relatively simple scenarios, precisely identifying and continuously tracking each target. In contrast, the high-density scenario of a busy waterway poses challenges such as dense target distribution, frequent occlusions, and intersecting trajectories. Although the model may have mismatches during operation (such as Scene 3 in Fig. \ref{figure:3s}), it can quickly recapture the temporal and spatial relationship characteristics between targets because of its use of dynamic graph modeling. In addition, when multiple targets cluster or intersect, the temporal characteristics extracted by the dynamic graph effectively differentiate between similar trajectories, thus preventing matching confusion.
\subsection{Ablation Analysis}
    To evaluate the contributions of the proposed TGA layer and STA block, we conducted an ablation study, as summarized in Table \ref{table:as}. Each component was removed individually to assess its impact on the model's performance. The absence of the TGA layer resulted in a significant performance degradation, with the average accuracy dropping to 37.67\%, underscoring its critical role in capturing local spatial relationships between vessels. Similarly, removing the STA block led to a decline in average accuracy to 56.04\%, highlighting its importance in modeling temporal variations, particularly in dynamic scenarios such as vessel turning or accelerating. When both components were integrated, the model achieved the highest average accuracy of 62.84\%, demonstrating the synergistic effect of combining the TGA layer and STA block in enhancing overall performance.
    \begin{figure}[t]
        \centering
        \setlength{\abovecaptionskip}{0.cm}
        \includegraphics[width=0.900\linewidth]{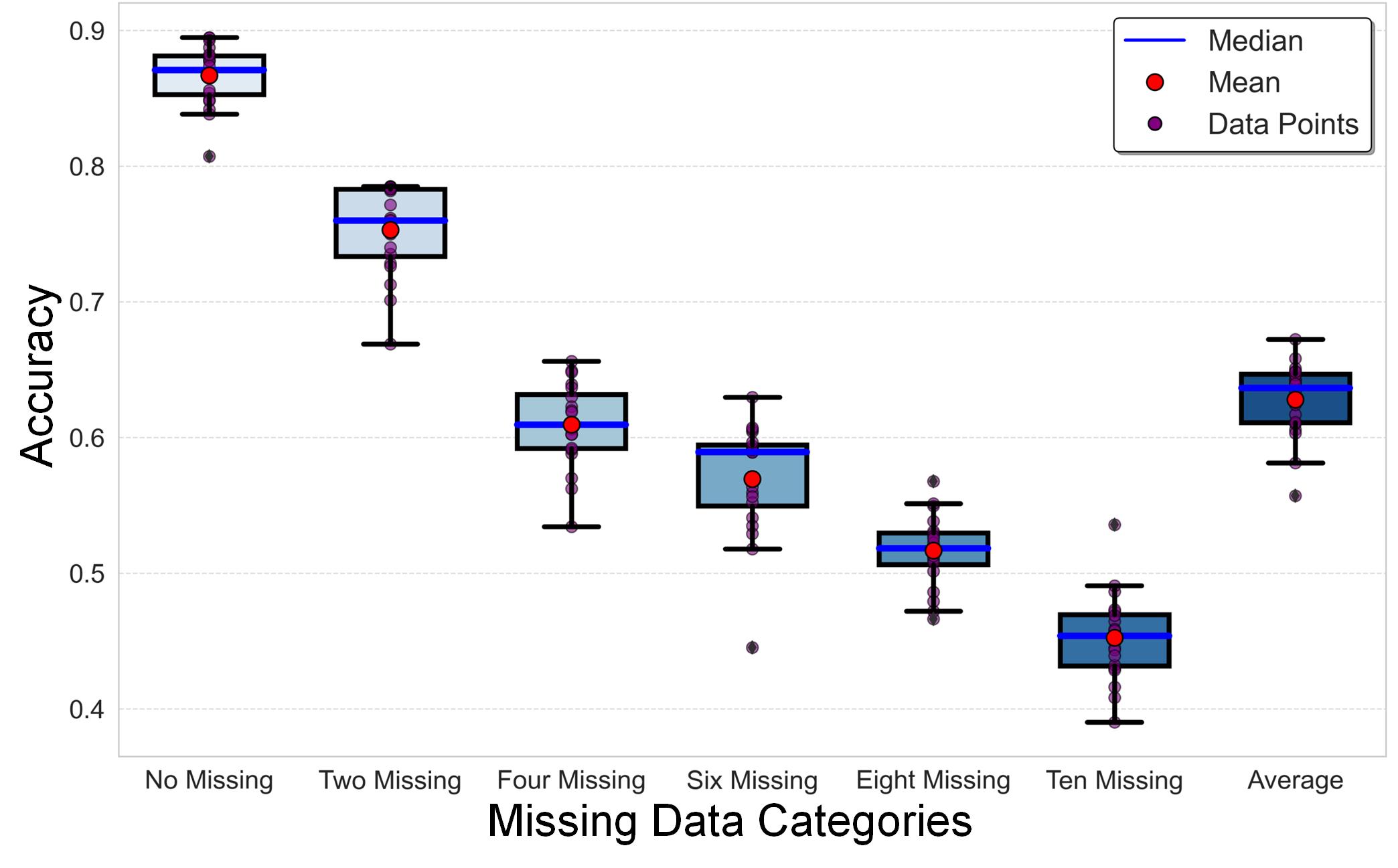}
        \caption{Stability analysis of our GMvA across missing data categories.}
        \label{figure:SA}
    \end{figure}
\subsection{Stability Analysis}
    This subsection systematically analyzes the stability of the GMvA method based on the results of 20 independent training experiments, focusing on the impact of data missingness on algorithm performance and its stability. As shown in Fig. \ref{figure:SA}, the results indicate that as the number of missing data increases, the accuracy of the method gradually declines: under conditions of no missing data, the average accuracy is the highest, remaining above 0.85; however, when the number of missing entries reaches 10, the average accuracy significantly drops to around 0.45, demonstrating the critical role of data integrity in method performance. Notably, even in the presence of missing data, the algorithm exhibits a narrow range of fluctuation in average accuracy across multiple training experiments, stabilizing between 0.35 and 0.41. This result highlights the stability and reliability of the GMvA method, indicating that its outputs are not random but consistently reproducible under varying training conditions.
    \setlength{\tabcolsep}{4.00pt}
    \begin{table}[t]
        \centering
        \footnotesize
        \caption{The computational complexity of different methods under different vessel density scenarios, including parameter size (unit: KB) and time consumption (unit: seconds).}
        \begin{tabular}{l|c|cccc}\hline
        \multirow{2}{*}{Methods} & \multirow{2}{*}{\makecell {Parameter \\ (KB)}} & \multicolumn{4}{c}{Time (second)}                       \\
        &                                 & \makecell {Low- \\ Density} & \makecell {Moderate- \\ Density}   & \makecell {High- \\ Density}   & Average \\\hline\hline
        ED          &  ---              & 0.0070      & 0.0142           & 0.0251       & 0.0154  \\
        CD          &  ---             & 0.0068      & 0.0135           & 0.0266       & 0.0156  \\
        PDF         &  ---              & 0.0069      & 0.0140           & 0.0272       & 0.0160  \\
        DTW         &  ---              & 0.0700      & 0.2787           & 0.6025       & 0.3170  \\
        GRU \cite{chung2014empirical}         & 3136           & 0.0343      & 0.0651           & 0.0913       & 0.0636  \\
        BiGRU \cite{cho2014learning}       & 6852           & 0.0375      & 0.0652           & 0.0980       & 0.0669  \\
        LSTM \cite{hochreiter1997long}        & 3910           & 0.0355      & 0.0636           & 0.0906       & 0.0632  \\
        BiLSTM \cite{schuster1997bidirectional}      & 7400           & 0.0385      & 0.0659           & 0.0965       & 0.0670  \\
        Transformer \cite{giuliari2021transformer} & 29982          & 0.0477      & 0.0810           & 0.1095       & 0.0794  \\
        GCN \cite{kipf2016semi}         & 2500           & 0.0355      & 0.0627           & 0.0876       & 0.0620  \\\hline
        GMvA        & 7435           & 0.0392      & 0.0664           & 0.0940       & 0.0665 \\\hline
        \end{tabular}\label{table:time}
    \end{table}
\subsection{Computational Time and Analysis}
    As shown in Table \ref{table:time}, our method demonstrates excellent computational efficiency and robustness across different vessel density scenarios. Our method is slightly higher in computational time than some traditional methods, mainly because the graph neural network (GNN) and multi-head attention mechanism are introduced into the network structure to capture complex temporal and spatial features, thereby improving the accuracy of matching. Although these modules increase the computational time, the average processing time is still in milliseconds to seconds. Furthermore, its ability to better handle both low- and high-density environments ensures its suitability for real-world applications such as waterway vessel monitoring and traffic management.
\section{Conclusion}\label{sec:conclusions}
    This work proposes a novel graph-based multi-vessel association (GMvA) method that effectively addresses the challenging multi-target association problem in maritime surveillance. By constructing dynamic graphs and leveraging GNN with spatiotemporal attention mechanisms, our method successfully captures complex vessel movement patterns from both CCTV and AIS data streams. The innovative integration of spatial convolution operations and temporal attention mechanisms, combined with an MLP-based uncertainty fusion module, enables robust feature learning and matching under challenging conditions. Our extensive experiments demonstrate that GMvA achieves significant performance improvements over traditional approaches, showing exceptional adaptability to varying traffic densities and strong robustness against data inconsistencies. The GMvA's ability to automatically learn optimal feature representations without relying on hand-crafted features or heuristic rules makes it particularly valuable for real-world maritime surveillance applications.

    However, extending GMvA to high-density scenarios remains challenging, as visual overlapping and increased computational complexity can hinder performance. Additionally, the reliance on precise data alignment between AIS and video sources also makes the method sensitive to missing data, noise, and preprocessing inconsistencies.

    Future work will focus on advanced graph construction and multi-modal fusion methods to improve multi-target association performance in dense traffic environments. Furthermore, integrating online learning with lightweight model optimization will enable real-time updates and efficient inference, ensuring robustness and scalability in complex and dynamic maritime applications. Additionally, we will explore incorporating radar data into the framework to further enhance multi-target association robustness, especially in challenging scenarios with visual occlusions or data inconsistencies.
\ifCLASSOPTIONcaptionsoff
\newpage
\fi
\bibliographystyle{IEEEtran}
\bibliography{ref.bib}
\tiny

\end{document}